# Improvement Tracking Dynamic Programming using Replication Function for Continuous Sign Language Recognition

S. Ildarabadi[#1], M. Ebrahimi[*2], H. R. Pourreza[#3]

[#]Sepahan Institute of high Education, Islamic Azad University Harand Branch, Islamic Azad University Mashad Branch

Iran

***Abstract:*** *In this paper we used a Replication Function (R. F.) for improvement tracking with dynamic programming. The R. F. transforms values of gray level [0 255] to [0 1]. The resulting images of R. F. are more striking and visible in skin regions. The R. F. improves Dynamic Programming (D. P.) in overlapping hand and face. Results show that Tracking Error Rate 11% and Average Tracked Distance 7% reduced.*

## 1. Introduction

The aim of tracking methods is to detect and follow one or several objects from a sequence of images. It can be seen as a kind of object detection in a series of similar images. In most cases, tracking methods are applied to video, which means that several images have to process. Therefore, the methods of object detection are not usually applicable, because they need too much computation time.

Many applications require processing in real-time. Furthermore, the knowledge about previous object positions can be used to predict and detect objects in following images.

Appearance-based features often use downscaled input images. Details in small images, like hand and face of the signer, are not exactly visible in the scaled images. It is possible to extract specific interest regions from video with tracking. The extracted regions path can be used as feature, It also useful for object position detection. In the field of sign language recognition tracking methods can be used to keep tracking of the hands and head. The tracked head and hands are used for visual and positional features computing.

Tracking methods are applied for many different tasks including gesture recognition, human movement tracking, face recognition, aerial surveillance and traffic supervision. A good overview on tracking methods, used in gesture and human movement recognition, can be found in [8]. Some of the widely methods are described briefly in [2].

The mean shift algorithm which tracks non-rigid objects based on visual features such as color and texture present on it. Statistical distributions are used to characterize interest objects. The algorithm tolerates partial occlusions of the tracked object, clutters, rotation in depth and changes in camera position.

The Continuously Adaptive Meanshift (Camshift) algorithm is a extension of the Meanshift algorithm that is able to deal with dynamically changing color probability distributions. [1] used it to track human faces in real-time.

In [9] the Conditional Density Propagation (Condensation) algorithm is presented. It is a model based method that is able to track objects in visual cluttered scenes.

An earlier version of tracking methods described in this work. It's applicable for sign language recognition. [4, 5, 6].

One of easily and rapidly tracking algorithm is Bounding-Box. It tries to recognize moving object positions with bounding box. This algorithm act upon images sequence and uses difference images for tracking the moving objects. This method tracks the moving pixels from one frame to another frame, because of prone error providing existence noise or whenever some moving objects exist in images. Also in this method background of images must be static and also object forwarder of another object in image. Difference images provide from submitting of sequence two frames.

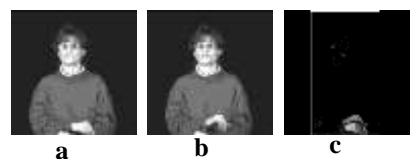

Fig1. Bounding-Box algorithm. (a), (b): Two sequence frame from database. (c): The result of Bounding-Box algorithm on (a) and (b) frames.

While have been correct hypothesis considered about background and moving object, this method provide good results. But in real applications is not correct hypothesis considered, thus not used this method. Figure 1 shows a sample.

## 2. Benchmark Databases: RWTH-Boston-Hands database-104

All databases presented Table 1 are used within the Sign Speak project and are either freely available or available on request. The SignSpeak1 project tackles the problem of automatic recognition and translation of continuous sign language [7]. The overall goal of the Sign Speak project is to develop a new vision-based technology for recognizing and translating continuous sign language (i.e. provide Video-to-Text technologies).

The RWTH-BOSTON-50 corpus was created for the task of isolated sign language recognition [11]. It has been used or nearest-neighbour leaving-one-out evaluation of isolated sign language words. About 1.5k frames in total are annotated and are freely available (cf. Table 1).

The RWTH-BOSTON-104 corpus has been used successfully for continuous sign language recognition experiments [5]. For the evaluation of hand tracking methods in sign language recognition systems, the database has been annotated with the signers' hand and head positions. More than 15k frames in total are annotated and are freely available (cf. Table 1). [7]

Table 1. Freely available tracking ground-truth annotations in sign language corpora for e.g. hand and face positions

| Corpus | Annotated Frames |
|---|---|
| RWTH-BOSTON-50 | 1450 |
| RWTH-BOSTON-104 | 15746 |
| Corpus-NGT | 7891 |
| ATIS-ISL | 5757 |
| OXFORD | 296 |

### 3. Tracking using dynamic programming

The tracking method introduced in [4, 10, and 11] is employed in this work. The used tracking algorithm prevents taking possibly wrong local decisions because the tracking is done at the end of a sequence by tracing back the decisions to reconstruct the best path. The tracking method can be seen as a two step procedure: in the first step, scores are calculated for each frame starting from the first, and in the second step, the globally optimal path is traced back from the last frame of the sequence to the first.

Step1. For each position $u = (i, j)$ in frame $x_t$ at time $t = 1,...T$ a score $q(t,u)$ is calculated, called the local score. The global score $C(t,u)$ is the total score for the best path until time $t$ which ends in position $u$. For each position $u$ in image $x_t$, the best predecessor is searched for among a set of possible predecessors from the scores $C(t-1,u')$; equation (1). This best predecessor is then stored in a table of back pointers $B(t,u)$ which is used for the trace back in Step 2; equation (2). This can be expressed in the following recursive equations:

$$C(t,u) = \max_{u' \in M(u)} \{C(t-1,u') - \tau(u',u)\} + q(t,u) \quad (1)$$

$$B(t,u) = \arg\max_{u' \in M(u)} \{C(t-1,u') - \tau(u',u)\} \quad (2)$$

, where $M(u)$ is the set of possible predecessors of point u and $\tau(u',u)$ is a jump-penalty, penalizing large movements.

Step2. The trace back process reconstructs the best path $u_1^T$ using the score table $C$ and the backpointer table $B$. Trace back starts from the last frame of the sequence at time $T$ using $c_T = \arg\max_u C(T,u)$. The best position at time $t-1$ is then obtained by $c_{t-1} = B(t,c_t)$. This process is iterated up to time $t =1$ to reconstruct the best path. Because each possible detecting center is not likely to produce a high score, pruning can be integrated into the dynamic programming tracking algorithm for speed-up. One possible way to track the dominant hand is to assume that this object is moving more than any other object in the sequence and to look at difference images where motion occurs to track these positions. Following this assumption, we use a motion information score function to calculate local scores using the first-order time derivative of an image. The local score can be calculated by a weighted sum over the absolute pixel values inside the detecting area. More details and further scoring functions are presented in [4, 10, and 11].

### 4. Improvement in dynamic programming method with gray level R. F.

At first, each video frame is passed from a fuzzy gray level R. F.. Figure (2) shows this function. It's triangle and mapped the gray level values (0 to 255) into (0 to 1).

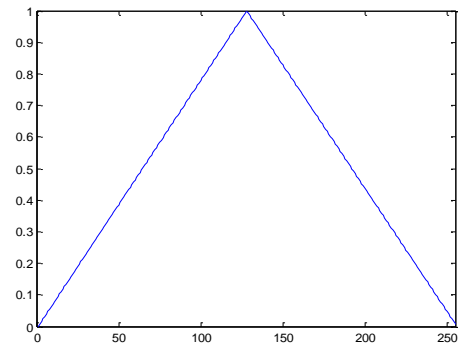

Fig2. The triangle gray level replication function for better detecting of mobile hand

Replication degree of each pixel, $p(i, j)$, obtain from equation (1). Some of passed images have been shown in figure (3).

A sequence of images $X_1^T = X_1,...,X_T$ is passed from R. F. and the results are $A_1^T = A_1,...,A_T$.

The score function, $q(u_{t-1}, u_t; A_{t-1}^t)$, is provided from subtracting of two sequential replicated frames; equation (3).

$$A_t^{'} = A^t - A^{t-1}$$
$$q(u_{t-1},u_t; A_{t-1}^t) = A_t^{'}[u] \quad (3)$$

That $u_{t-1}$ and $u_t$ are the positions of hand in $A^{t-1}$ and $A^t$ frames.

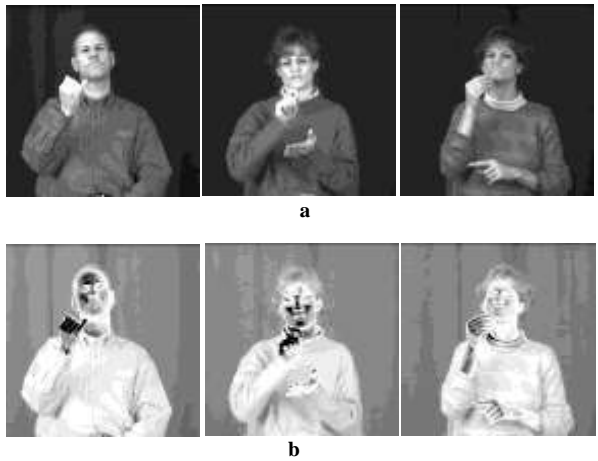

Fig3. Row (a): three instance of database images. Row (b): images of first row after passed from gray level R. F.. (Equation 8)

An example has been shown in figure (4) and (5). Figure (4) is two consecutive frames of a sentence of database, with the difference image of them. Figure (5) is same of figure (4) but they affected with gray level R. F..

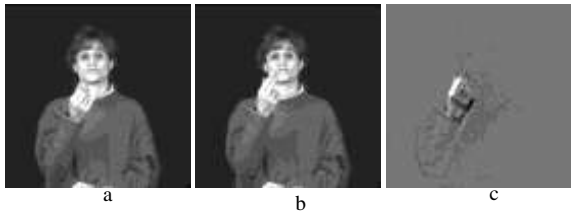

Fig4. The difference of two sequential frames. (a), (b): two sequential frames. (c): subtracted of two frames.

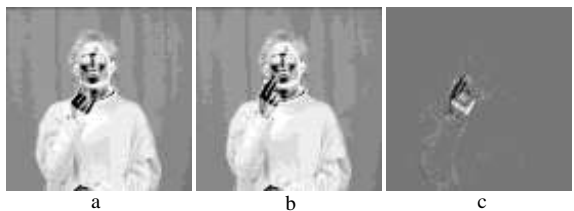

Fig5. The differences of two sequential frames of figure (4) after passing them from R. F..

The function $\tau$ is penalty of jumping. In here, Euclidean distance has been used as penalty function; equation (8).

$$\tau(i,j) = \alpha\sqrt{i^2 + j^2} \qquad (8)$$

, where $\alpha$ is weighting factor for the jumping penalty. We propose that the maximum the speed of moving hand from one frame to another isn't more than n pixels, therefore $\tau$ is $(2n+1) \times (2n+1)$ matrix. The $\tau$ has been shown in figure 6 for n=5 and $\alpha = 1$

| 7.071 | 6.403 | 5.831 | 5.385 | 5.099 | 5 | 5.099 | 5.385 | 5.831 | 6.403 | 7.071 |
|---|---|---|---|---|---|---|---|---|---|---|
| 6.403 | 5.657 | 5 | 4.472 | 4.123 | 4 | 4.123 | 4.472 | 5 | 5.657 | 6.403 |
| 5.831 | 5 | 4.243 | 3.606 | 3.162 | 3 | 3.162 | 3.606 | 4.243 | 5 | 5.831 |
| 5.385 | 4.472 | 3.606 | 2.828 | 2.236 | 2 | 2.236 | 2.828 | 3.606 | 4.472 | 5.385 |
| 5.099 | 4.123 | 3.162 | 2.236 | 1.414 | 1 | 1.414 | 2.236 | 3.162 | 4.123 | 5.099 |
| 5 | 4 | 3 | 2 | 1 | 0 | 1 | 2 | 3 | 4 | 5 |
| 5.099 | 4.123 | 3.162 | 2.236 | 1.414 | 1 | 1.414 | 2.236 | 3.162 | 4.123 | 5.099 |
| 5.385 | 4.472 | 3.606 | 2.828 | 2.236 | 2 | 2.236 | 2.828 | 3.606 | 4.472 | 5.385 |
| 5.831 | 5 | 4.243 | 3.606 | 3.162 | 3 | 3.162 | 3.606 | 4.243 | 5 | 5.831 |

Fig6. The function $\tau$ with $\alpha = 1$ and $n = 5$.

Function $C(t,u)$, (Equation 5), is introduced, which gives the best score for the path at time $t$, ending in position $u$.

Function $C(t,u)$ is defined recursively. The score $C(t,u)$ is calculated for each time step $t$ and each position $u$ successively, starting from $t = 1$, $C(1,u_1) = q(u_1, u_2, A_1^2)$ yielding a table of scores. The maximization does not need to consider all predecessor positions of position $u$, but a limited set of predecessor positions $M(u)$; equation (6). This limitation avoids large distances between consecutive object positions (additional to the jump penalty) and decreases computation time.

$$C(t,u) = \max_{u' \in M(u)} \left\{ C(t-1,u') - \tau(u',u) + q(u',u; A_{t-1}^t) \right\} \qquad (5)$$

$$M(u_{i,j}) = C(x+i, y+j), \quad -n/2 \le i,j \le n/2 \qquad (6)$$

To reconstruct the best path, a table of back-pointers $B(t,u)$ is needed, which stores the best predecessor position for each time step $t$ and each position $u$; equation (7):

$$B(t,u) = \arg\max_{u' \in M(u)} \left\{ C(t-1,u') - \tau(u',u) + q(u',u; A_{t-1}^t) \right\} \qquad (7)$$

Function $C(t,u)$, catches the location of target, means the right hand, and gives the location of it at previous frames.

The best path is traced back as follows:
1. Search best position at last frame and select a pixel, $u_T$, from it using Otsu's method (part 3-1).
2. Repeat for $t = T-1$ down to $t = 1$, means $u_{t-1} = B(t, u_t)$.

We used the RWTH-BOSTON-104 database [3].

### 4.1 Image threshold for Segmentation of right hand using Otsu's method, at last frame

Otsu's method chooses the threshold to minimize the intraclass variance of the black and white pixels.

We used Otsu's method twice to segment the hand area. At first time, person area segmented from background frame (figure 7). At second time, skin of person segmented from person area (figure 8).

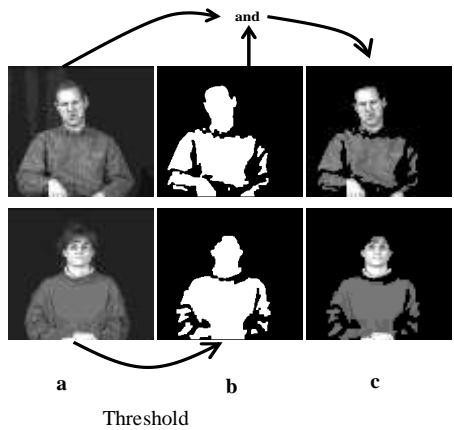

Fig7. Segmentation of person area using Otsu's method. (a): two image from database. (b): segmentation of image using Otsu's threshold. (c): logical and of image (a) and (b).

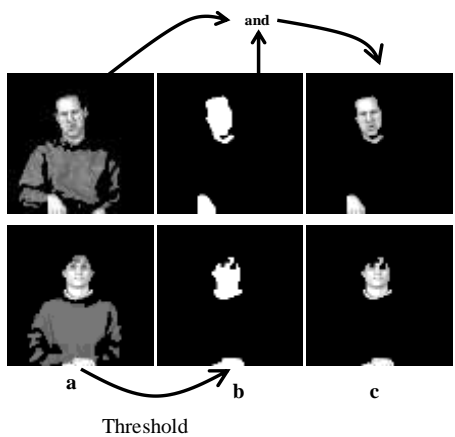

Fig8. Segmentation of skin area using Otsu's method. (a): two image of figure 7 part (c). (b): segmentation of image using Otsu's threshold. (c): logical and of image (a) and (b).

We must select one pixel from right hand area. We use it for tracking of hand at prior frames (figure 9). In last frame, the right hand is at bottom and left.

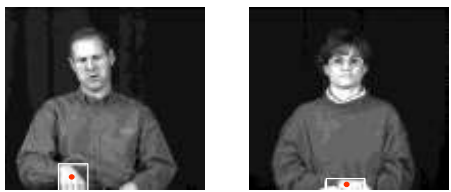

Fig9. One pixel is selected from right hand area.

### 4.2 Hand Tracking

The four parameters that have been used were:
a- The collection of priority points (equation (9))

$$M(i,j) = \{(i+i', j+j') : -J \leq i', j' \leq J\} \text{ with } J = 10.$$

b- A window with 20*20 sizes.
c- Penalty weight of jumping ($\lambda_\tau$) equal to 0.1.
d- Using of subtracted images as score function.
All parameters were fixed in both of Dynamic Programming, D. P., and improved D. P. The result of D. P. algorithm has been seen in figures (10) and (11).

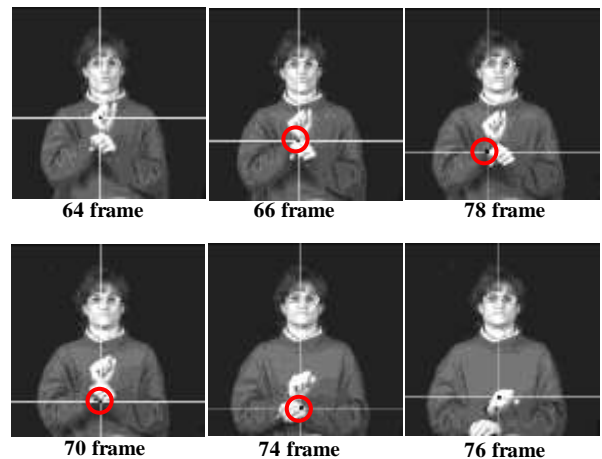

Fig10. Results of D. P., The mistake detecting of left hand instead of right hand in 66, 68, 70,74 frames

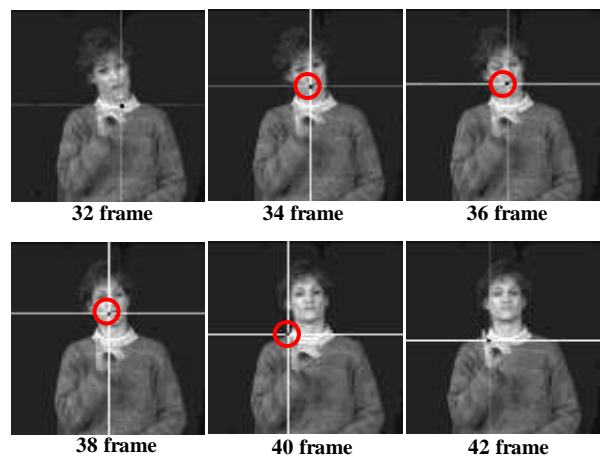

Fig11. Results of D. P., The mistake detecting of head instead of right hand in 34, 36, 38 frames

The result of improved algorithm has been seen in figs. (12) and (13).

We see that proposed algorithm tracked the hand better than ancient algorithm. When we compare the figures (10) and (11) with the figures (12) and (13), we will find the effect of R. F. on prior method.

It means, when a hand covered the face, the D. P. algorithm without pass in frames cannot track the hand has error.

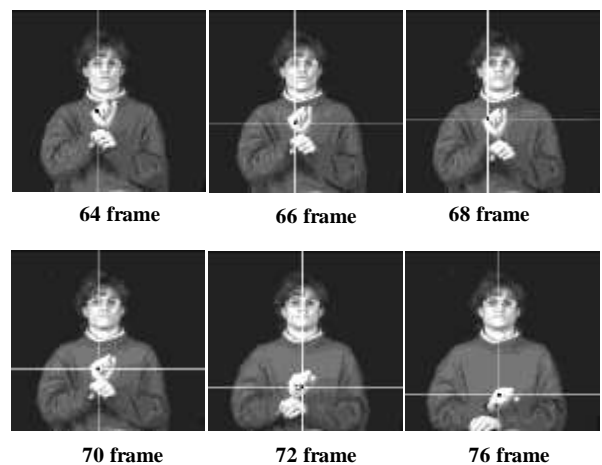

Fig12. The Results of improved D. P. without error

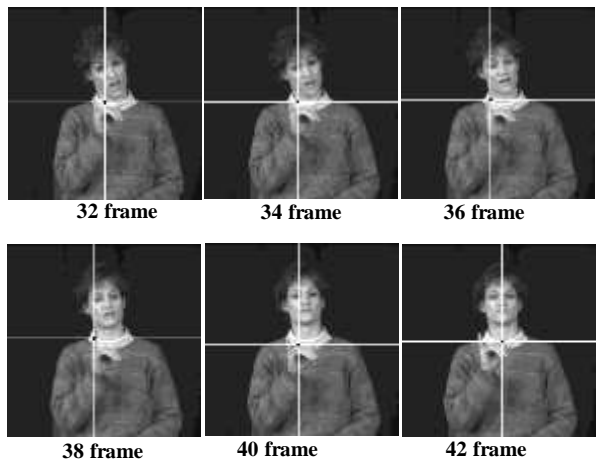

32 frame    34 frame    36 frame
38 frame    40 frame    42 frame

Fig13. The Results of improved D. P. without error
In proposed tracking algorithm, every image, at first, passed with R. F. then the D. P. uses it.

## 5. Results

Table2 shows our results.

### 5.1 Efficiency measurement criterion in tracking algorithms.

To comparing different tracking methods, we introduce two measurements criterion:

### 5.1.1 Tracking Error Rate

Tracking Error Rate, TER, shows the number of windows which have tracked the target truly; equation (8).

We define $\tau$, as a parameter for more exact surveying. It is shortest distance between the center of tracked window and the center of nearest same window which has covered the target completely, for a sequence of images. In those, Euclidean distance between tracked situation and marked situation has been equal to $\tau$ or higher than it.

$$TER = \frac{1}{T}\sum_{t=1}^{T}\delta_\tau(u_t, \hat{u}_t) \quad (8)$$

with $\delta_\tau(u,v) := \begin{cases} 0 & \|u-v\| < \tau \\ 1 & \text{otherwise} \end{cases}$

### 5-1-2- Average Tracked Distance

Average Tracked Distance, ATD, the average distance between tracked situation and marked situation; equation (10).

$$\mu_d = \frac{1}{T}\sum_{t=1}^{T}\|u_t - \hat{u}_t\| \quad (9)$$

These two priorities don't distinguish the positive errors from the negative errors.

Table2 shows our results. The R. F. improves D. P. in overlapping hand and face. Results show that TER 11.27% and $\mu_d$ 7.12% reduced.

Table2. Evaluation of D. P. [11] and improved D. P.

|  | (%)TER | | (%) $\mu_d$ |
|---|---|---|---|
|  | $\tau = 15$ | $\tau = 20$ |  |
| D. P. | 19.83 | 9.43 | 12.92 |
| Improved D. P. | 8.56 | 2.23 | 5.80 |

### 5.2 Recommendation

There are three basic stages in sign language recognition systems: Hand Tracking, Feature Extraction and Sign Recognition. The detecting section has important role in these systems.

D. P. method is one of the best tracking algorithms. It has done better than others same meanshift [2]. In this paper we try to improve D.P. algorithm by means of gray level R. F.: we passed video frames into the R. F.. Gray level digits (0 to 255) transfer to (0 to 1). Then subtraction of these frames has been used as score function in D.P. method for better tracking.

The other important point in here is that we must use all frames of sentence to tracking the hand. So this algorithm acts offline. To establishing online systems, we must collect the small tracking parts and combine them to find the meaning.